%% file: manuscript.tex
\documentclass[10pt,twocolumn,letterpaper]{article}

\input{macros}

\input{math_commands.tex}

\usepackage[pagebackref=true,breaklinks=true,letterpaper=true,colorlinks,bookmarks=false]{hyperref}
\cvprfinalcopy 

\def\cvprPaperID{3766} 

\ifcvprfinal\fi
\begin{document}

\title{End-to-end Learning for Graph Decomposition}

\author{Jie Song$^{1}$ \quad \quad Bjoern Andres$^{3,4,5}$ \quad \quad Michael Black$^{2}$ \quad \quad Otmar Hilliges$^1$ \quad \quad Siyu Tang$^{2,5}$\\
\small $^{1}$ ETH Zurich \quad \quad
\small $^{2}$ MPI for Intelligent Systems  \quad \quad
\small $^{3}$ MPI for Informatics \quad \quad
\small $^{4}$ Bosch Center for AI \quad \quad
\small $^{5}$ University of T\"ubingen 
}

\maketitle

\begin{abstract}
We propose a novel end-to-end trainable framework for the graph decomposition problem. The minimum cost multicut problem is first converted to an unconstrained binary cubic formulation where cycle consistency constraints are incorporated into the objective function. The new optimization problem can be viewed as a Conditional Random Field (CRF) in which the random variables are associated with the binary edge labels of the initial graph and the hard constraints are introduced in the CRF as high-order potentials. The parameters of a standard Neural Network and the fully differentiable CRF are optimized in an end-to-end manner.
Furthermore, our method utilizes the cycle constraints as meta-supervisory signals during the learning of the deep feature representations by taking the dependencies between the output random variables into account.
We present analyses of the end-to-end learned representations, showing the impact of the joint training, on the task of clustering images of MNIST.  
We also validate the effectiveness of our approach both for the feature learning and the final clustering on the challenging task of real-world multi-person pose estimation. 
\end{abstract}

\input{chapters/introduction}

\input{chapters/related_work}
\input{chapters/method}
\input{chapters/application}
\input{chapters/conclusion}

\newpage
{\small
\bibliographystyle{ieee}
\bibliography{cvpr-2019-crf-pose}
}

\end{document}

%% file: macros.tex
\usepackage{cvpr}
\usepackage{times}
\usepackage{epsfig}
\usepackage{graphicx}
\usepackage{amsmath}
\usepackage{amssymb}

\usepackage{algorithmic}
\usepackage{algorithm}
\usepackage{rotating}
\usepackage{diagbox}
\usepackage{balance}

\usepackage[utf8]{inputenc} 
\usepackage[T1]{fontenc}    
\usepackage{url}            
\usepackage{booktabs}       
\usepackage{amsfonts}       
\usepackage{nicefrac}       
\usepackage{microtype}      
\usepackage{graphicx}       
\usepackage{paralist}
\usepackage{xspace}
\usepackage{comment}
\usepackage{subfig}

\usepackage{amssymb,amsmath}
\newcommand{\ccycles}{\textnormal{cc}}
\newcommand{\myparagraph}[1]{\vspace{0.1em}\noindent\textbf{#1}}
\usepackage{tikz}
    \definecolor{mycolor1}{HTML}{009900}
    \definecolor{mycolor2}{HTML}{990000}
    \definecolor{beamer@blendedblue}{HTML}{000000}
    \usetikzlibrary{decorations.pathmorphing}
    \usetikzlibrary{patterns}
    \tikzstyle{mynode}=[circle, draw=black, fill=white, inner sep=0pt, minimum width=1ex]
    \tikzset{every picture/.append style={baseline,scale=1.1}}



%% file: math_commands.tex
\usepackage{amsmath,amsfonts,bm}

\def\eqref#1{~\ref{#1}}

\def\1{\bm{1}}

\DeclareMathAlphabet{\mathsfit}{\encodingdefault}{\sfdefault}{m}{sl}
\SetMathAlphabet{\mathsfit}{bold}{\encodingdefault}{\sfdefault}{bx}{n}



%% file: chapters/introduction.tex
\section{Introduction}
Many computer vision problems, e.g. multi-person pose estimation \cite{pishchulin2016deepcut}, instance segmentation \cite{kirillov-2017}, and multi-target tracking \cite{tang-2015}, can be viewed as optimization problems, where decompositions of a graph are the feasible solutions.
For example, in multi-person pose estimation, a graph $G=(V,E)$ can be constructed where the nodes $V$ correspond to body joint detections and the edges $E$ connect the detections that hypothetically indicate the same person \cite{pishchulin2016deepcut}. 
Partitioning the detections that describe the same person into the same connected component with respect to the graph $G$ is a Minimum Cost Multicut Problem \cite{Chopra:1993:PP:161127.161133,Bansal2004}, with respect to a linear objective function. 

It has several appealing properties: First, in contrast to other balanced cut problems \cite{Shi:2000:NCI:351581.351611}, it does not favor one decomposition over another. Instead of relying on a fixed number of graph components or biasing them by the problem definition, in this formulation the number of decompositions is determined by the solution in an unbiased fashion. 
Second, it is straightforward to utilize this optimization problem in practice: 
for many vision tasks, an input graph can be easily constructed and the cost of the incident nodes being in distinct components can be obtained robustly from some Deep Neural Networks, e.g. \cite{insafutdinov-2017,kirillov-2017}.



%
%

By far, the most common way of applying the minimum cost multicut problem to vision tasks is to employ a multi-stage pipeline \cite{kirillov-2017,pishchulin2016deepcut,insafutdinov-2016,tang-2017-multiple}. Briefly speaking, 
first, the task dependent detections and the affinity measures between the detections are obtained by two separately trained networks. Second, the coefficients of the objective function are constructed based on the output of the networks and third, the optimization is performed independently on top of the detection graph by either branch and bound algorithms \cite{pishchulin2016deepcut,tang-2015} or heuristic greedy search algorithms \cite{cao2017realtime}.


While being straightforward, a notable caveat of the multi-stage approach is that the deep networks are learned locally without utilizing the knowledge of how to perform the graph decomposition globally. The dependencies among the optimization variables 
are not considered during the training of the deep feature representations.
Notably, several works have shown that combining graphical models such as Conditional Random Fields (CRFs) with deep learning to train feature representations can result in remarkable performance gains~\cite{Zheng:2015:CRF:2919332.2919659,tompson2014joint}. 
It is, however, an open question how to develop a learning algorithm that can learn a better deep feature representation when taking into account the variables' global dependencies defined by a general graph decomposition problem, such as the minimum cost multicut problem. 

Motivated by this question, we propose a novel end-to-end trainable framework for the joint learning of feature representation and graph decomposition problem. 
We first convert the minimum cost multicut problem to an unconstrained binary cubic problem to incorporate the hard consistency constraints into the objective function.
The appealing property of this new optimization problem is that it can be viewed as a conditional random field (CRF).
The random variables of the CRF are associated with the binary edge labels of the initial graph, and the hard constraints are introduced as high-order potentials in the CRF. 
We further propose an end-to-end learnable framework that consists of a standard Convolutional Neural Network (CNN) as a front end and a fully differentiable CRF with the high-order potentials.
The advantages of the proposed framework are: 
\begin{inparaenum}[(i)]
\item The parameters of the CRF and the weights of the front end CNN are optimized jointly during the training of the full network via backpropagation. Such joint training facilitates a learnable balance between the unary potentials and high-order potentials that enforce the validity of the edge labeling, which leads to a better decomposition. 
%
\item The cycle inequalities, encoded by the high-order potentials, serve as supervision signals during learning of the deep feature representations. This meta-supervision from the global consistency constraints is complementary to the direct local supervision (standard CNN training) in the way that it teaches the network how to behave by taking the dependencies between the output random variables into account.
\end{inparaenum} 

In experiments, we first present analyses on the task of clustering MNIST (\cite{Lecun98gradient-basedlearning}) images , showing it is beneficial for the feature learning by enforcing the global consistency constraints.
We then demonstrate the proposed model on the challenging task of multi-person pose estimation in unconstrained images. Our results suggest the effectiveness of the end-to-end learning framework in terms of better feature learning, cycle constraint validity, tighter confidence of the marginal estimates and final pose estimation performance. 






%

%

%% file: chapters/related_work.tex
\section{Related Work}
\myparagraph{The minimum cost multicut problem.} 
The multicut problem has been explored for various computer vision tasks \cite{pishchulin2016deepcut,insafutdinov-2017, tang-2015,kirillov-2017,keuper-2015a,levinkov2017comparative}. \cite{keuper-2015a} applies it to motion segmentation, where pixel-wise motion trajectories are clustered into individual moving objects. In \cite{pishchulin2016deepcut,insafutdinov-2016}, a joint node and edge labeling problem is proposed to model the multi-person pose estimation task. In \cite{tang-2015,tang-2017-multiple}, the multi-target tracking task is formulated as a graph decomposition problem.
Meanwhile, many algorithms for efficiently solving the minimum cost multicut
problem have been proposed \cite{7298973,Kappes,Kappes-2,NIPS2011_4406,5878,Yarkony}. \cite{7298973} proposes a correlation clustering fusion method which iteratively improve the current solution by a fusion operation.
\cite{Yarkony} relies on column generation to combine feasible solutions
of subproblems into successively better solution in planar graphs.
\cite{swoboda-2017} proposes a dual decomposition and linear program
relaxation algorithm which alternates between message passing and separation
of cycle and odd-wheel inequalities efficiently. 
There are also algorithms that integrate optimization problems as layers into network architectures for end-to-end training~\cite{pmlr-v70-amos17a,Schulter2017DeepNF,FrossardU18,Zanfir_2018_CVPR}.
\cite{Zanfir_2018_CVPR} constructs different matrix layers of their computation graph, the analytic derivatives are obtained by matrix backpropagation.  \cite{Schulter2017DeepNF} proposes an end-to-end learning framework for the cost functions of the network flow problem and multi-object tracking is the target application in the paper. \cite{pmlr-v70-amos17a} proposes a general method of integrating quadratic programs with deep networks. Due the cubic complexity in the number of constraints, it is an open question whether this method can be used for the complex vision tasks. 
To the best of our knowledge, ours is the first work that introduces an end-to-end learnable framework for the multicut formulation. 

\myparagraph{Learning deep structured model.} 
Several works have been proposed to jointly learn the feature representations and the structural dependency between the variables of interest~\cite{Chen:2015:LDS:3045118.3045308,Arnab2016HigherOC,newell2017associative, chu2016crf}. \cite{Chen:2015:LDS:3045118.3045308} proposes a learning framework to jointly estimate the deep representations and the parameters of their Markov random field model.
\cite{Zheng:2015:CRF:2919332.2919659} proposes to formulate the mean field iterations as recurrent neural network layers, and ~\cite{Arnab2016HigherOC} further extends \cite{Zheng:2015:CRF:2919332.2919659} to include their object detection and superpixel potentials for the task of semantic segmentation.
\cite{chu2016crf} proposes a CRF-CNN model to incorporate the structural information into the hidden feature layers of their CNN.
The goal of our work is to design an end-to-end learning framework for the minimum cost multicut problem. Although the mean field inference used here does not guarantee producing a feasible graph decomposition, it can be effectively integrated into a CNN to facilitate the desired joint training.

\myparagraph{Human pose estimation.}
Recent deep neural network based methods have made great progress on human pose estimation in natural images in particular for the single person case~\cite{tompson2014joint,newell2016stacked,wei2016convolutional,Nie_2018_CVPR,Peng_2018_CVPR,Luo_2018_CVPR}. As for a more general case where multiple people are present in images, previous work can mainly be grouped into either top-down or bottom-up categories. Top-down approaches first detect individual people and then predict each person’s pose~\cite{fang2017rmpe,papandreou2017towards,he2017mask}. One of the challenges for top-down approaches is that they make detection decisions at a very early stage, which is fragile and prone to false negatives.  Bottom-up approaches directly detect individual body joints and then associate them with individual people~\cite{cao2017realtime,insafutdinov-2016,insafutdinov-2017,newell2017associative}. 
In~\cite{pishchulin2016deepcut,cao2017realtime}, the body joint detections and the affinity measures between the detections are first trained by deep networks, then the association is performed independently either by branch and bound algorithms~\cite{pishchulin2016deepcut} or by heuristic greedy search algorithms~\cite{cao2017realtime}.  One potential advantage over top-down approaches is that the decision making of detections (typically non-maximum suppression is deployed) is performed at lower levels (joints) rather than at the highest level (person).
Our work is also related to \cite{newell2017associative}. The difference is that our method focuses on end-to-end learning the graph decompostion problem, in \cite{newell2017associative} the associations are trained by predicting person IDs directly along with joint detections.


%% file: chapters/method.tex
\section{Optimization Problem}
\label{sec:optimization}
\subsection{Minimum Cost Multicut Problem}
The minimum cost multicut problem \cite{Chopra:1993:PP:161127.161133,Bansal2004} is a constrained binary linear program w.r.t.~a graph $G = (V,E)$ and a cost funciton $c: E \to \mathbb{R}$:
\vspace{-0.2em}
\small
\begin{align}
\min_{y \in \{0,1\}^E} \label{mc-objective} \quad
    & \sum_{e \in E} c_e \, y_e\\
\textnormal{subject~to} \label{mc-constraints} \quad
    & \forall C \in \ccycles(G) \,
        \forall e \in C: \quad
            y_e \leq \sum_{e' \in C \setminus \{e\}}
                y_{e'}
\enspace.
\end{align}
\normalsize
Here, the optimization variables $y\in \{0,1\}^E$ correspond to a binary labeling of the edges $E$. $y_e =1 $ indicates that the edge $e$ is cut. In other words, the nodes $v$ and $w$ connected by edge $e$ are in distinct components of $G$. 
$\ccycles(G)$ denotes the set of all chord-less cycles of $G$. The cycle constraints in Eq.~\ref{mc-constraints} define the feasible edge labellings, which relate one-to-one to the decompositions of the graph $G$. 
A toy example is illustrated in Fig.~\ref{figure:mc-crf}: (a) shows an example graph $G$; (b) is a valid decomposition of $G$; and (c) shows an invalid solution that violates the cycle inequalities (Eq.~\ref{mc-constraints}). 
The cost function $c: E \to \mathbb{R}$ is characterized by model parameters $\theta$.
In previous work \cite{pishchulin2016deepcut,insafutdinov-2016,insafutdinov-2017}, the cost function is defined as
$\log\frac{1-p_e}{p_e}$, where $p_e$ denotes the probability of $y_e$ being cut. Given a feature $f_e$ on the edge $e$, $p_e$ takes a logistic form: 
$ \frac{1}{1 + \exp(-\langle \theta, f_e \rangle)}$.
The maximal probable model parameters $\theta$ are then obtained by maximum likelihood estimation on training data. $f_e$ can be  attained via some deep feature representations extracted from a separately trained deep network. For example, in \cite{insafutdinov-2017} and \cite{tang-2017-multiple}, $f_e$ is obtained from a convolutional neural network and a Siamese network respectively.  

At the heart of this work lie the following research questions: first, how to jointly optimize the model parameters $\theta$ and the weights of the underlying deep neural network for the graph decomposition problem? 
Second, how to utilize the cycle consistency constraints as supervision signal and to capture the dependencies between the output random variables during training? In the following, we present our end-to-end learnable framework which provides solutions to these research questions.
%
%

\subsection{Unconstrained Binary Cubic Problem}
Our first observation is that the minimum cost multicut problem can be equivalently stated as an unconstrained binary multilinear program with a large enough constant $C\in \mathbb{N}$
\small
\begin{align}
\min_{y \in \{0,1\}^E} \quad
     \sum_{e \in E} c_e \, y_e
        + C \sum_{C \in \ccycles(G)} \sum_{e \in C} y_e
            \prod_{e' \in C \setminus \{e\}}
                (1 - y_{e'})
\enspace .
\label{eq:crf-mc}
\end{align}
\normalsize
In the special case where $G$ is complete, every 3-cycle is chordless.
Thus, Eq.~\eqref{eq:crf-mc} specializes to the binary \emph{cubic} problem as described in Eq.~\eqref{eq:crf-mc-3} where  $\bar y_{vw} := 1 - y_{vw}$.
\small
\begin{align}
\min_{y \in \{0,1\}^E} \quad
      & \sum_{e \in E} c_e \, y_e 
        +  C \sum_{\{u, v, w\} \in \tbinom{V}{3}} 
        (y_{uv} \bar y_{vw} \bar y_{uw} \nonumber \\
             + & \bar y_{uv} y_{vw} \bar y_{uw} 
             + \bar y_{uv} \bar y_{vw} y_{uw})
\enspace .
\label{eq:crf-mc-3}
\end{align}
\normalsize
An invalid cycle inequality, e.g. in Fig.~\ref{figure:mc-crf}(c) where $y_{vw} =1, y_{uw} = y_{uv} = 0$ and  $\bar y_{uv} y_{vw} \bar y_{uw} = 1$, contributes a constant value $C$ into the objective (Eq.~\ref{eq:crf-mc-3}). By setting $C$ to be large enough, the second terms in Eq.~\ref{eq:crf-mc-3} equal to 0, and the cycle consistent constraints defined in Eq.~\ref{mc-constraints} are satisfied.
\input{figures/mc-crf}

\subsection{Multicut as Conditional Random Fields}
\label{sec:CRF}
Our second observation is that the unconstrained binary cubic problem (Eq.~\ref{eq:crf-mc-3}) can be expressed by a Conditional Random Field with unary potentials that are defined on each edge variable and high-order potentials that are defined on every three edge variables.
More specifically, we define a random field over the variables $\mathbf{X} = (X_1,X_2 \cdots,X_{|E|})$ that we want to predict. $\mathbf{I}$ is the observation, e.g. an image. We associate each random variable $x_i$ with an edge variable $y_e$ in Eq.~\ref{eq:crf-mc-3}, and the random variable  $x_i$ takes a value from a label set $\{0,1\}$. 
Then the optimization problem (Eq.~\ref{eq:crf-mc-3}) can be expressed as the following CRF model:

\begin{align}
E(\mathbf{x}|\mathbf{I}) = \sum_{i}\psi_{i}^{U}(x_{i}) + \sum_{c}\psi_{c}^{Cycle}(\mathbf{x}_{c})
\label{eq:crf-general}
\end{align}
where $E(\mathbf{x}|\mathbf{I})$ is the energy associated with a configuration $\mathbf{x}$ conditioned on the observation $\mathbf{I}$. 
Our goal is to obtain a labeling with minimal energy, namely $\mathbf{\hat{x}} \in \mathrm{argmin}_\mathbf{x}E(\mathbf{x}|\mathbf{I})$. 
Such a labeling is the maximum a posteriori (MAP) solution of the Gibbs distribution $P(\mathbf{X}=\mathbf{x}|\mathbf{I}) = \frac{1}{Z(\mathbf{I})} \exp{-E(\mathbf{x}|\mathbf{I})}$ defined by the energy $E(\mathbf{x}|\mathbf{I})$, where $Z(\mathbf{I})$ is the partition function.

The unary potentials $\psi_{i}^{U}(x_{i})$ correspond to the first terms in Eq.~\ref{eq:crf-mc-3}, measuring the inverse likelihood of an edge being cut. 
It can take arbitrary forms. 
As shown in Sec.~\ref{sec:end-to-end learning}, in case of multi-person pose estimation, $\psi_{i}^{U}(x_{i})$ utilizes the output of a state-of-the-art CNN ~\cite{cao2017realtime}.

The high-order terms $\psi_{c}^{Cycle}(\mathbf{x}_{c})$ are one of the key contributions of this work. They are introduced to model the cycle inequalities (Eq.\eqref{mc-constraints}) in the minimum cost multicut problem and correspond to the second terms in Eq.~\ref{eq:crf-mc-3}. Each high-order potential associates a cost to a cycle in the initial graph. The primary idea is that, for every cycle in the graph, a high cost will incur if the current edge labellings in the cycle violate the cycle consistency constraint. 

\paragraph{Pattern-based Potentials.}
There is a finite set of valid edge labellings for 3-cycles in the graph. Fig.~\ref{figure:mc-crf} illustrates a simple graph and examples of valid (1-1-0) and invalid (1-0-0) edge labellings.
To assign high/low cost for the invalid/valid cycles, we utilize the pattern-based potentials proposed in \cite{5206846}.
\small
\begin{align}
\psi_{c}^{Cycle}(\mathbf{x}_{c})=\begin{cases}
               \gamma_{\mathbf{x}_{c}} & \text{if}\ \mathbf{x}_{c} \in \mathcal{P}_{c}\\
               \gamma_{max} & \text{otherwise},
            \end{cases}
\label{eq:pattern-potential}
\end{align}
\normalsize

where $\mathcal{P}_{c}$ is the set of recognized label configurations for the clique, namely, valid cycles in the initial graph. We assign a cost $ \gamma_{\mathbf{x}_{c}} $ to each of them. $\gamma_{max}$ is then assigned to all the invalid label configurations for the clique, namely, invalid cycles in the initial graph. 

Given the proposed potentials, minimizing the energy of the proposed CRF model (\ref{eq:crf-general}) is then equivalent to minimizing the optimization problem defined in Eq.~\eqref{eq:crf-mc-3}.



\paragraph{Inference.}
We resort to mean field inference to minimize the energy defined in Eq.~\ref{eq:crf-general},
which has been formulated as a Recurrent Neural Network and integrated into a CNN framework \cite{Zheng:2015:CRF:2919332.2919659}.
For the mean field inference, an alternative distribution $Q(\mathbf{x})$ defined over the random variables is introduced to minimize the KL-divergence between $Q(\mathbf{x})$ and the true distribution  $P(\mathbf{x})$. The general mean field update follows \cite{Koller:2009:PGM:1795555}:
\small
\begin{align}
Q_{i}(x_i = l) = \frac{1}{Z_i} \exp\{-\sum_{c\in C}\sum_{\{ \mathbf{x}_c | x_i= l \}} Q_{c-i}(\mathbf{x}_{c-i})\dotfill \psi_c(\mathbf{x}_c)\} .
\label{eq:mean-field-update}
\end{align}
\normalsize
Here $\mathbf{x}_c$ is a configuration of all the variables in the clique $c$ and
$\mathbf{x}_{c-i}$ is a configuration of all the variables in the clique $c$ except $x_i$.
Given the definition of the pattern-based potential in Eq.~\ref{eq:pattern-potential}, 
The mean field updates for our CRF model can be derived from the work of \cite{Vineet2014} as:
\footnotesize
\begin{align}
Q^{t}_{i}(x_i = l) = & \frac{1}{Z_i} \exp\{-\sum_{c\in C}( \sum_{p\in \mathcal{P}_{c|x_i=l}}( \prod_{j\in c,j\neq i} Q^{t-1}_j(x_j = p_j))\gamma_{p} \nonumber \\
& + \gamma_{max}(1-\sum_{p\in \mathcal{P}_{c|x_i=l}}( \prod_{j\in c,j\neq i} Q^{t-1}_j(x_j = p_j)))))\}
\label{eq:mean-field-update-ours}
\end{align}
\normalsize
where $x_j$ represents a random variable in the clique $c$ apart from $x_i$,
$\mathcal{P}_{c|i=l}$ is the subset of $\mathcal{P}_{c}$ where $x_i = l$. 
$t$ denotes the $t^{th}$ iteration of the mean field inference. 
Assume $L$ is the value of a loss function defined on the result obtained by the mean filed inference,
Eq.~\ref{eq:mean-field-update-ours} allows us to backpropagate the error $\frac{\partial L}{\partial Q}$
to the input $\mathbf{x}$ as well as the parameters $\gamma_{\mathbf{x}_{c}}$ and $\gamma_{max}$. 

Note that after the mean field inferences, it is not guaranteed to obtain a valid graph decomposition, as the mean field inference enforces the validity of the cycle consistency but does not guarantee that all the hard constraints (E.g. \ref{mc-constraints}) are fulfilled. Therefore in practice, we resort to some fast heuristics (E.g. \cite{keuper-2015a}) to return a feasible graph decomposition after the mean field inferences.

\paragraph{Learning.}
Although the mean field update (Eq.~\ref{eq:mean-field-update-ours}) does not guarantee that all the hard constraints (E.g. \ref{mc-constraints}) are fulfilled, it allows us to backpropagate the error signals, which facilitates an end-to-end learning mechanism. More specifically, we are now able to jointly optimize the deep feature representation and the parameters for performing the partitioning of the graph, by reformulating the original optimization problem to the CRF model.
Concretely, the following parameters can be jointly learned by the proposed model via backpropagation:

\begin{itemize}
\item[--] $W$ which are the weights of the front-end deep neural network
\item[--] $\theta$ which characterizes the cost function $c: E \to \mathbb{R}$ in the minimum cost multicut problem
\item[--] $\gamma_{\mathbf{x}_{c}}$ and $\gamma_{max}$ that are introduced by the high-order potentials of the CRF model.
\end{itemize}
By the joint training, the dependencies between the optimization variables are incorporated into the learning for a better deep feature representation via the proposed high-order potentials. 

\subsection{ A Toy Example on MNIST}
To understand how the proposed end-to-end learning model integrates the dependencies between the output random variables during training, we consider a simple task that clusters images of hand-written digit (MNIST \cite{Lecun98gradient-basedlearning}) {\em without} specifying the number of clusters.
This problem can be formulated as a minimum cost multicut problem (Eq.~\ref{mc-objective}-\ref{mc-constraints}) that is defined on a fully connected graph. The nodes of the graph indicate the digit images and edges connect the images that hypothetically indicate the same digit.
Using this simple task, we discuss how the following two approaches learn the deep feature representation to associate the images.

\myparagraph{Approach I: Train a Siamese network.} A straight forward way to obtain the similarity measures between any two images is to train a Siamese network which takes a pair of images as input and produces a probabilistic estimation of whether the image pair indicates the same or different digits. We use the architecture of LeNet~\cite{lecun1998gradient} which is commonly used on digit classification tasks. Fig.~\ref{figure:mnist-example} shows two example results.
In Fig.~\ref{figure:mnist-example}(a), the probabilities for the top/left pair and left/right pair being the same digit are 0.94 and 0.85 respectively, which are correctly estimated. But for the top/right pair, it is 0.47, likely due to the high intra-class variations.
Similarly for the example in Fig.~\ref{figure:mnist-example}(b), the probability for the top/right pair being the same digit is incorrectly estimated.
When we partition these digits into clusters, the incorrectly estimated similarity measures introduce invalid cycles. Now the question is whether we can utilize such cycle constraints to learn a better Siamase network, which could produce more robust and consistent similarity measures.


\myparagraph{Approach II: Train the Siamese network and the CRF jointly.}
In this approach, we aim to train the Siamese network by taking the cycle consistency constraints into account. We resort to the proposed model, where we convert the partitioning problem to the energy minimization problem defined on the CRF (Eq.~\ref{eq:crf-general}). 
Specifically, we add a stack of customized CRF layers that perform the iterative mean field updates with high-order potentials on top of the Siamese network (the customized CRF layers are introduced by \cite{Zheng:2015:CRF:2919332.2919659} and the details will be described in Sec.~\ref{sec:mean_Field_update}). Now we are able to train the Siamese network and the CRF model jointly. As to the examples in Fig~\ref{figure:mnist-example} (a),  the probability of the top/right pair indicating the same digit is increased to 0.56, using the end-to-end learned Siamese network and it is further improved to 0.61 after the mean-field updates with the jointly learned CRF parameters. 
As to the overall performance, the accuracy of similarity measures produced directly by the Siamese network is increased from 91.5\% to 93.2\%. The corresponding final clustering accuracy is increased from 94.1\%   to 95.9\%.

Despite being conceptually simple, the LeNet-based Siamese network and the customized CRF layers are able to be jointly learned and produce more robust and consistent results on the task of clustering MNIST digits.
The next open question is how to design an end-to-end learnable framework for real-world challenging vision tasks that rely on clustering.


\input{figures/MNIST.tex}

%% file: figures/mc-crf.tex
\begin{figure}[t]
\centering
\scriptsize
\tikzstyle{mynode} = [circle, draw=black, inner sep=0ex, minimum size=2ex]
\tikzstyle{mynode_rec} = [rectangle, draw=black, inner sep=0ex, minimum size=2ex, fill=black]
\begin{minipage}{0.25\textwidth}
    \centering

    \begin{tikzpicture}[scale=0.5]
    \node[mynode, label=below:$u$] at (0, 0) (v0) {};
    \node[mynode, label=above:$v$] at (1, 1.5) (v1) {};
    \node[mynode, label=below:$w$] at (2, 0) (v2) {};
    \draw (v0) -- (v1);
    \draw (v1) -- (v2);
    \draw (v0) -- (v2);
    \node at (1,-0.8) {(a)};
    
    \node[mynode, label=below:$u$] at (3, 0) (v3) {};
    \node[mynode, label=above:$v$] at (4, 1.5) (v4) {};
    \node[mynode, label=below:$w$] at (5, 0) (v5) {};
    \draw (v3) -- (v4);
    \draw [dotted, thick](v4) -- (v5);
    \draw [dotted, thick](v3) -- (v5);
 	\node at (4,-0.8) {(b)};
 
    \node[mynode, label=below:$u$] at (6, 0) (v6) {};
    \node[mynode, label=above:$v$] at (7, 1.5) (v7) {};
    \node[mynode, label=below:$w$] at (8, 0) (v8) {};
    \draw (v6) -- (v7);
    \draw [dotted, thick](v7) -- (v8);
    \draw (v6) -- (v8);
    \node at (7,-0.9) {(c)};
    
    \end{tikzpicture}
\end{minipage}
\begin{minipage}{0.15\textwidth}
    \centering

    \begin{tikzpicture}[scale=0.5]
    \node[mynode] at (0, 0) (v0) {};
    \node[mynode] at (1, 0) (v1) {};
    \node[mynode] at (2, 0) (v2) {};
    \node[mynode_rec] at (0, -1) (v3) {};
    \node[mynode_rec] at (1, -1) (v4) {};
    \node[mynode_rec] at (2, -1) (v5) {};
    \node[mynode_rec, label=above:$ $] at (1, 1) (v6) {};
    \draw (v0) -- (v3);
    \draw (v1) -- (v4);
    \draw (v2) -- (v5);
    \draw (v0) -- (v6);
    \draw (v1) -- (v6);
    \draw (v2) -- (v6);
    \node at (1,-1.6) {(d) };
    \node at (0, -0.2) {}; 
    \end{tikzpicture}
\end{minipage}
\caption{We illustrate a graph $G$ in (a); a feasible solution and an infeasible solution are shown in (b) and (c) respectively; the factor graph of the CRF model of the graph $G$ is in (d).}
\label{figure:mc-crf}
\end{figure}
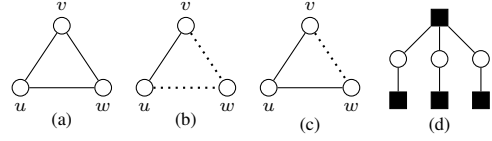

%% file: figures/MNIST.tex
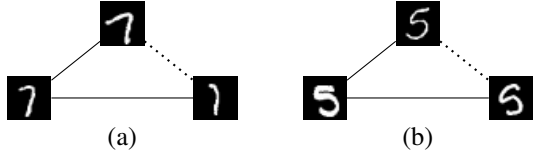
\begin{figure}[t]
\centering
\begin{minipage}{0.22\textwidth}
    \centering
    \begin{tikzpicture}[scale=0.45]
    \node[inner sep=0pt] (7_1) at (0, 0)
    {\includegraphics[width =.15\textwidth]{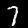}};
    \node[inner sep=0pt] (7_2) at (2.5, 2)
    {\includegraphics[width =.15\textwidth]{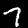}};
    \node[inner sep=0pt] (7_3) at (5, 0)
    {\includegraphics[width =.15\textwidth]{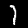}};
    \draw (7_1) -- (7_2);
    \draw  (7_1) -- (7_3);
    \draw [dotted, thick](7_2) -- (7_3);
    \node at (2.5,-1.1) {(a)};
    \end{tikzpicture}
\end{minipage}
\begin{minipage}{0.22\textwidth}
    \centering
    \begin{tikzpicture}[scale=0.45]
    \node[inner sep=0pt] (7_1) at (0, 0)
    {\includegraphics[width =.15\textwidth]{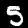}};
    \node[inner sep=0pt] (7_2) at (2.5, 2)
    {\includegraphics[width =.15\textwidth]{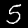}};
    \node[inner sep=0pt] (7_3) at (5, 0)
    {\includegraphics[width =.15\textwidth]{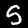}};
    \draw (7_1) -- (7_2);
    \draw [dotted, thick] (7_2) -- (7_3);
    \draw (7_1) -- (7_3);
    \node at (2.5,-1.1) {(b)};
    \end{tikzpicture}
\end{minipage}
\vspace{-1em}
\caption{Examples of inconsistent edge labels produced by a stand alone Siamese network on the MNIST digits. 
}
\label{figure:mnist-example}
\end{figure}

%% file: chapters/application.tex
\section{End-to-End Learning for Multi-person Pose Estimation}
\label{sec:end-to-end learning}

In this section, we further design an end-to-end learnable framework for the challenging multi-person pose estimation task. Our network consists of four parts: 1) a front end CNN that outputs feature representations (Sec.\ref{sec:cnn_to_unary}); 2) two fully connected layers to convert the features to the unary potentials (Sec.\ref{sec:cnn_to_unary}); 3) a stack of customized layers that perform the iterative mean field updates (Sec.\ref{sec:mean_Field_update}) and 4) the loss layer that is on top of the mean field iteration (Sec.\ref{sec:loss}).
We choose multi-person pose estimation as case study because this task is considered to be one of the  fundamental problems in understanding people in natural images.
Recent work \cite{he2017mask,pishchulin2016deepcut,cao2017realtime,insafutdinov-2017} has made significant progress by the driving force of deep feature learning. 
For instance, the work proposed by Cao et al.~\cite{cao2017realtime} presents a powerful deep neural network to learn feature representation for body joints and limbs, followed by a fast heuristic matching algorithm to associate body joints to individual pose. 
Given the performance on public benchmarks of \cite{cao2017realtime}, 
in the following, we utilize their pre-trained network as the front end CNN. Our model is complementary to  \cite{cao2017realtime} in the way that our focus is the joint optimization of the deep feature learning and the detection association.  

\subsection{From CNN to Unary Potentials}
\label{sec:cnn_to_unary}
\myparagraph{Network Architecture.} The network proposed in \cite{cao2017realtime} has two separate branches after sharing the same basic convolutional layers: one branch predicts the confidence maps for 14 body joints and the other branch estimates a set of part affinity fields, which encode joint to joint relations.
The part field is a 2D vector field. More specifically, each pixel in the affinity field is associated with an estimated 2D vector that encodes the direction pointing from one joint to the other. In \cite{cao2017realtime}, the part fields are implemented only for pairs of joints that follow the kinematic tree of the human body, e.g. left elbow to left hand. However, in order to incorporate high order potentials among neighboring joints, we train the model to also capture the feature between jump connections, e.g. shoulder to wrist.

%

\myparagraph{Graph Construction.} Given an input image, we first obtain the body joint candidates from the detection confidence maps. For each type of the joint, we keep multiple detection hypotheses even for those that are in close proximity. A detection graph is then conducted in the way that we insert edges for pairs of hypotheses that describe the same type of body joint, and for pairs of hypotheses between two different joints. Note that, although the constructed graph is not fully connected, every chordless cycle in the graph consists of only three edges. 

\myparagraph{Edge Feature.} The key to the robust graph decomposition is a reliable feature representation on the edges to indicate whether the corresponding joint detections belong to the same/different person. For the edges that connect the detection hypotheses of different body types, we use the corresponding part field estimation. More specifically, we compute the inner product between the unit vector defined by the direction of the edge and vectors that are estimated by the part field. We collect 10 values by uniformly sampling along the line segment defined by the edge. These values form the feature $f_e$ for the corresponding edge.
For the edges that connect the detection hypotheses of the same joint type, we simply use the euclidean distance between the detection as the feature. A better way would be to design another branch for the network \cite{cao2017realtime} to predict whether the two detections of the same joint type describe the same person; we leave this for future work.

\myparagraph{The Unary $\psi^{U}$.} It is straightforward to construct the unary potentials $\psi_{i}^{U}(x_{i})$ (Eq.~\ref{eq:crf-general}) from the edge feature $f_e$. We incorporate two  fully connected layers to encode the feature to classify if an edge is cut, namely, the two corresponding  joints belong to different persons. 
As described in Sec.~\ref{sec:CRF}, during training, we can obtain the error signal from the mean field updates to learn the parameters of the newly introduced fully connected layers and the front end CNN that produces the edge feature.

\subsection{Mean Field Updates}
\label{sec:mean_Field_update}
Zheng et al.~\cite{Zheng:2015:CRF:2919332.2919659} propose to formulate the mean field iteration as recurrent neural network layers, and ~\cite{Arnab2016HigherOC} further extend it to include high-order object detection and superpixel potentials for the task of semantic segmentation. In this work, we follow their framework with the modification of incorporating our pattern-based potentials.
The goal of the mean field iterations is to update the marginal distribution  $Q^{t}_{i}(x_i = l)$. For initialization, $Q^{1}_{i}(x_i = l) = \frac{1}{Z_i} \exp\{-\psi_{i}^{U}(x_{i}=l)\}$, where $Z_i = \sum_l \exp\{-\psi_{i}^{U},(x_{i}=l)\}$ is performed. This is equivalent to applying a soft-max function over the negative unary energy across all the possible labels for each link. This operation does not include any parameters and the error can be back-propagated to the front end convolutional or fully connected layers where the unary potentials come from. Once the marginal has been initialized, we compute the high order potentials based on Eq.~\eqref{eq:mean-field-update-ours}. Specifically, the valid cliques in $\mathcal{P}_{c}$ are 0-0-0, 1-1-1 and 1-1-0, while the non-valid cliques are 0-0-1, where 1 indicates that the corresponding edge is cut. This operation is differentiable with respect to the parameters $\gamma_{\mathbf{x}_{c}}$ and $\gamma_{max}$ introduced in Eq.~\eqref{eq:mean-field-update-ours}, allowing us to optimize them via backpropagation. The errors can also flow back to  $Q^{1}(X)$. Once the high order potential is obtained, it is summed up with the unary potential and then the sum is normalized via the soft-max function to generate the new marginal for the next iteration. Multiple mean-field iterations can be efficiently implemented by stacking this basic operation. During the inference, as the mean field inference does not guarantee a feasible solution to the original optimization problem, we use the fast heuristic proposed in \cite{cao2017realtime} as an additional step to come back to the feasible set.

\subsection{Loss and Training}
\label{sec:loss}
During training, we first train the joint confidence maps and part affinity field maps with  a standard $L2$ loss as described in~\cite{cao2017realtime}. Once the basic features are learned, the next step is to train the unary with the softmax loss function. This is performed in an on-the-fly manner, which means the detection hypotheses for the body joints are estimated and then the links between the hypotheses are also established during training time. Their ground-truth labels are also generated online at the same time. The final step is to train the parameters of the CRF with high order potentials with a softmax loss function in an end-to-end manner along with the basic convolutional and fully connected layers.

\begin{table}[t]
\begin{center}
\resizebox{\linewidth}{!}{
\begin{tabular}{l|cccccccc}
\hline
& H-N &N-S  & S-E &E-W &S-Hi &Hi-K &K-A  & Mean \\
\hline
 origin &  0.755 & 0.656 & 0.662 & 0.558& 0.679& 0.593 & 0.611& 0.635 \\
\hline
 Iter 1&   0.783 & 0.692 & 0.688 & 0.579& 0.711& 0.628 & 0.639& 0.659 \\
 Iter 2&   0.805 & 0.707 & 0.715 & 0.603& 0.726& 0.649 & 0.651& 0.671 \\
 Iter 3&   0.807 & 0.712 & 0.716 & 0.608& 0.723& 0.646 & 0.653& 0.674 \\

\hline
\end{tabular}
}
\end{center}
\vspace{-2em}
\caption{\textbf{Marginal distribution updates.} Numbers represent evolution of the marginal probabilities along with the mean-field iterations for different type of limbs.} 
\label{table:marginal}
\end{table}

\begin{table}
\begin{center}
\footnotesize
\resizebox{\linewidth}{!}{
\begin{tabular}{l|c|c|c|c|c}
\hline
& H-N-S  & S-E-W  &N-LH-RH &H-K-A & Mean \\
\hline
 origin &  1.68 & 3.40 & 1.41 & 3.83& 2.60  \\
 \hline
 Iter 1&   1.12 & 2.79 & 1.06 & 3.17& 2.04  \\
 Iter 2&   1.01 & 2.58 & 0.89 & 2.82& 1.81 \\
 Iter 3&   0.96 & 2.47 & 0.87 & 2.79& 1.76 \\
\hline
\end{tabular}
}
\end{center}
\vspace{-2em}
\caption{\textbf{Ratio of non valid cycle.} Numbers (\%) represent the ratio of non valid cycle for four different types of cliques that are defined for adjacent body joints.}
\label{table:non valid cycle}
\end{table}

\subsection{Experiments}

\myparagraph{Dataset.}
We use the MPII Human Pose dataset which consists of about 25k images and contains around 40k total annotated people. 
There is a training and test split with 3844 and 1758 groups of people respectively. 
We conduct ablation experiments on a held out validation set. 
During testing, no information about the number of people or the scales of individual is provided. 
For the final association evaluation, we deploy the evaluation metric proposed by~\cite{pishchulin2016deepcut}, calculating the average precision of the joint detections for all the people in the images.
 In the following experiments, we use shortcuts for body joints (Head-H, Neck-N, Shoulder-S, Elbow-E, Wrist-W, Hip-Hi, Knee-K, Ankle-A).

\myparagraph{Implementation Details.}
The front-end CNN architecture has several stacked fully convolutional layers with an input size of 368x368 as described in \cite{cao2017realtime}. We train the basic CNN using a batch size of 12 with a learning rate of 1e-4. 
For training the CRF parameters, the learning rate is 1e-5. 
The whole architecture is implemented in Caffe~\cite{jia2014caffe}.

\myparagraph{Effectiveness of the CRF Inference.}
To demonstrate the effectiveness of our proposed mean-field layers approximating the CRF inference, we evaluate the evolution of the marginal distribution for the random variables $X$.
For pose estimation, each variable $X_i$ in the CRF represents a link between two body joints. 
As seen in Tab.~\ref{table:marginal}, 7 different types of limbs are depicted. 
The numbers are the average marginal probabilities for those links with the ground truth of not being cut. 
It measures how confident a link is supposed to be associated. In other words, the confidence that two joints belong to the same person.
As shown in the table, the marginal distributions of all the limbs benefit from high order potentials even for very challenging combinations, e.g. Elbow-Wrist and Knee-Ankle. After three iterations of inference, the update converges and we fix this setting for further experiments.

\myparagraph{Validity of the Cycle Constraints.}
Another important measurement for our proposed model is to check the ratio change of non-valid cycles after the mean field iterations. As mentioned in Sec.~\ref{sec:end-to-end learning}, the type of non-valid 3-clique is link-link-cut. We can see from Tab.~\ref{table:non valid cycle} that, with the CRF inference, the ratios of non-valid cycles decrease, indicating the effectiveness of the high order potential.

\begin{figure*}[h]
	\includegraphics[width=\textwidth]{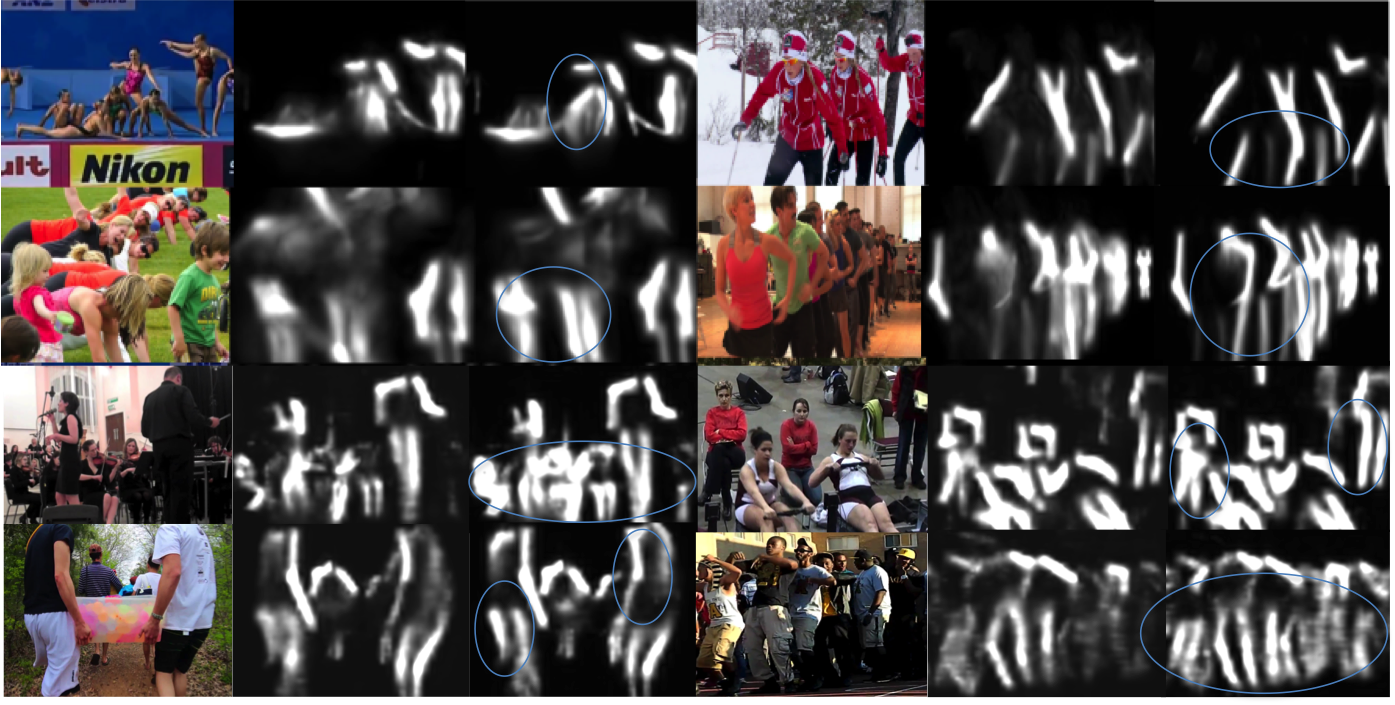}
    \caption{\textbf{Feature learning comparison.} Left: input image; Middle: part field map learned locally, without considering the cycle consistency; Right: part field map learned with the cycle consistency. The right samples clearly show sharper and more accurate confidence maps.}
    \label{fig:part-field map}
\end{figure*}

\myparagraph{Benefit of End-to-End Learning on Feature Representation.}
One of the key advantages of training the CNN and CRF jointly is to obtain a better feature representation. 
%
We illustrate it by directly visualizing the part field feature maps before and after the mean field inference.
As shown in Fig.~\ref{fig:part-field map}, the confidence maps in general get sharper and cleaner, particularly for images with heavy occlusions; e.g.~in the second image in the second row, the limbs of the partially occluded people become more distinguishable, suggesting a notable improvement in the feature learning for the challenging cases. 
This is in line with one of the assumptions of this work: 
the deep features needs additional supervision signals from the high-order terms, particularly for challenging cases.

%
%
%

\begin{figure*}[h]
	\includegraphics[width=1\textwidth]{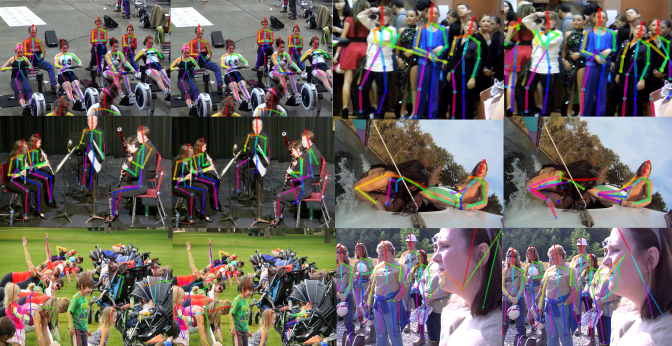}
    \caption{\textbf{Qualitative Results.}  Left: association  without CRF inference; Right: association after inference. \textbf{First row}, obvious wrong connections are corrected by inference. In the \textbf{second row} occluded people are separated. The samples in the \textbf{last row} are failure cases.}
    \label{fig:association}
\end{figure*}

\myparagraph{Return to a Feasible Solution.} After the CRF inference, we do not obtain a valid graph decomposition directly. Some heuristics (either the greedy search~\cite{cao2017realtime} or the KL heuristic~\cite{insafutdinov-2017}) are required to generate a valid decomposition efficiently. We evaluate these two heuristics with three different settings for each: 1) only front-end CNN and full connected layers (unary); 2) trained CRF on top of front-end CNN and fully connected layers (unary and CRF); 3) end-to-end finetuning of the whole network ( end-to-end finetuning).
Tab.~\ref{table:validation} shows the analysis on the validation set and we can see the advantage of the end-to-end strategy over the offline training of CRF as a
post-processing method. As illustrated in Fig.~\ref{fig:association}, the improvements are mainly achieved on the challenging cases with heavy occlusion, which benefit from modeling the high-order dependency among the variables of interest.

\myparagraph{Comparison With Others.}
We test the proposed method on the MPII Human Pose dataset and compare with other methods.
The result are shown in Tab.~\ref{table:c}.
Our end-to-end method achieves 76.1 mAP, which is comparable with other state-of-the-art methods.
Note that the method proposed in \cite{newell2017associative} uses a single-person pose estimator to refine the final result, and \cite{fang2017rmpe} is a top-down method where a Faster R-CNN \cite{Ren:2015:FRT:2969239.2969250} person detector is utilized.

\begin{table} [t]
\centering
\resizebox{\linewidth}{!}{
\begin{tabular}{l|ccccccc|c}
\hline
 Method& Head & Shou &Elbo &Wris &Hip &Knee &Ankl &Mean \\
\hline

unary (KL)& 88.55  &  83.98 & 71.43  & 60.97  & 73.44  & 65.25 & 56.66 & 71.32 \\
unary and CRF (KL) & 89.04  & 84.36  & 72.01  & 61.39  & 73.68  &66.75  & 58.11 & 71.96 \\
end-to-end (KL) &  89.38 & 84.72 & 72.49  & 61.96  &  74.05 &  66.85 & 58.46 & 72.57 \\
unary (greedy)& 91.30 & 86.14   &  73.69 & 62.84  &  73.40 & 66.43 & 58.73   & 73.21 \\

unary and CRF (greedy)& 91.31 & 86.47  & 74.60  & 64.01 & 73.69 &  66.98& 59.45 & 73.78 \\
end-to-end (greedy) & 91.50 & 86.90   &  74.89 & 64.61  &  73.97 & 67.38 & 59.91 & 74.36 \\
\hline
\end{tabular}
}
\vspace{-1em}
\caption{Multi-person pose estimation result on the validation set. }
\label{table:validation}
\end{table}

\begin{table}
\centering
\resizebox{\linewidth}{!}{
\begin{tabular}{l|ccccccc|c}
\hline
 Method& Head & Shou &Elbo &Wris &Hip &Knee &Ankl &Mean \\
\hline
Insafutdinov et al., \cite{insafutdinov-2016}& 78.4  & 72.5  & 60.2  & 51.0  & 57.2  & 52.0 & 45.4 & 59.5 \\
pishchulin et al., \cite{pishchulin2016deepcut}& 89.4  & 84.5  & 70.4  & 59.3  & 68.9  & 62.7 & 54.6 & 70.0 \\
Insafutdinov et al., \cite{insafutdinov-2017}& 88.8  & 87.0  & 75.9  & 64.9  & 74.2  & 68.8 & 60.5 & 74.3 \\
Cao et al., \cite{cao2017realtime}& 91.2  & 87.6  & 77.7  & 66.8  & 75.4  & 68.9 & 61.7 & 75.6 \\
\hline
Fang et al., \cite{fang2017rmpe}& 88.4  & 86.5  & 78.6  & 70.4  & 74.4  & 73.0 & 65.8 & 76.7 \\
Newell et al., \cite{newell2017associative}& 92.1  & 89.3  & 78.9  & 69.8  & 76.2  & 71.6 & 64.7 & 77.5 \\
Nie et al., \cite{nie2018pose}& 92.2  & 89.7  & 82.1  & 74.4  & 78.6  & 76.4 & 69.3 & 80.4 \\

\hline
Our Method & 91.4 & 87.8  & 78.0  & 67.2  &76.5 & 69.3 &62.2 & 76.1 \\
\hline
\end{tabular}
}
\vspace{-1em}
\caption{Comparison with the state-of-the-art on the MPII Human Pose dataset.}
\label{table:c}
\end{table}

%% file: chapters/conclusion.tex
\section{Conclusion, Limitation and Future Work}
In this work, our goal is to answer the following research questions: (1) how to jointly optimize the model parameters and  the  weights  of  the  underlying  deep  neural  network for the graph decomposition problem?  (2) how to use the cycle consistency as a supervision signal to capture the dependencies of the output random variables during training?  To that end, we propose to convert the minimum cost multicut problem to an energy minimization problem defined on a CRF. 
The hard constraints of the multicut problem are formulated as high-order potentials of the CRF whose parameters are learnable. 
We perform analyses on the task of clustering digit images and multi-person pose estimation. The results validate the potential of our method and show improvement both for the feature learning and the final clustering task.


Although, as we show in this work, the proposed learning method for the multicut problem has several strong points, there are still some limitations. First, with the proposed mean field update, we can jointly learn the front end deep networks and the parameters of the graph decomposition. However, the hard constraints in the optimization problem are not guaranteed to be satisfied. Therefore during testing, we resort to efficient heuristic solvers to return a feasible graph decomposition. 
Second, we show notable improvement on the feature learning and validity of the cycle inequality for the multi-person pose estimation task, but the final performance gain on pose association does not support us to outperform the state-of-the-art top-down methods. One reason is that current evaluation metric favors top-down methods. Another reason is that our end-to-end training only operates on the part affinity field, not on the body joint detections, which is crucial for the final result. To include the body joint detections in the end-to-end training pipeline is a practical future direction. Nevertheless, We think that this work adds an important primitive to the toolbox of the graph decomposition problem and opens up many avenues for future research.

\noindent {\bf Acknowledgments.}
 S. T. acknowledges funding by the German Research Foundation (DFG CRC 1233). {\bf Disclosure} MJB has received research gift funds from Intel, Nvidia, Adobe, Facebook, and Amazon. While MJB is a part-time employee of Amazon, his research was performed solely at, and funded solely by, MPI.